\documentclass[accepted]{uai2023} 

\usepackage[american]{babel}

\usepackage{natbib} 
\usepackage{mathtools} 
\usepackage{booktabs} 
\usepackage{tikz} 

\usepackage{amsfonts} 
\usepackage{enumerate}
\usepackage{algorithm2e}
\usepackage{tablefootnote}

\title{Deep learning and MCMC with aggVAE for shifting administrative boundaries: mapping malaria prevalence in Kenya}

\author[1,5]{\href{mailto:<elizaveta.p.semenova@gmail.com>?Subject=Your UAI 2023 paper}{Elizaveta Semenova}{}}
\author[2,5]{Swapnil Mishra}
\author[3,5]{Samir Bhatt}
\author[1,5]{Seth Flaxman}
\author[4,5]{H Juliette T Unwin}
\affil[1]{%
    Department of Computer Science\\
    University of Oxford
}
\affil[2]{%
    Saw Swee Hock School of Public Health and Institute of Data Science\\
    National University of Singapore and NUHS
}
\affil[3]{%
    School of Public Health, University of Copenhagen; School of Public Health, Imperial College London
  }
\affil[4]{%
    School of Public Health, Imperial College London; Department of Mathematics, University of Bristol
  }
\affil[5]{%
    Machine Learning and Global Health Network (www.MLGH.net)

  }

\begin{document}
\twocolumn[
  \begin{@twocolumnfalse}
    \maketitle

    \vspace*{1cm} 
  \end{@twocolumnfalse}
]

\section*{Abstract}
Model-based disease mapping remains a fundamental policy-informing tool in the fields of public health and disease surveillance. Hierarchical Bayesian models have emerged as the state-of-the-art approach for disease mapping since they are able to both capture structure in the data and robustly characterise uncertainty. When working with areal data, e.g.~aggregates at the administrative unit level such as district or province, current models rely on the adjacency structure of areal units to account for spatial correlations and perform shrinkage. The goal of disease surveillance systems is to track disease outcomes over time. This task is especially challenging in crisis situations which often lead to redrawn administrative boundaries, meaning that data collected before and after the crisis are no longer directly comparable. Moreover, the adjacency-based approach ignores the continuous nature of spatial processes and cannot solve the change-of-support problem, i.e.~when estimates are required to be produced at different administrative levels or levels of aggregation. We present a novel, practical, and easy to implement solution to solve these problems relying on a methodology combining deep generative modelling and fully Bayesian inference: we build on the recently proposed PriorVAE method able to encode spatial priors over small areas with variational autoencoders by encoding aggregates over administrative units. We map malaria prevalence in Kenya, a country in which administrative boundaries changed in 2010. 

\section{Introduction}\label{sec:intro}

\begin{figure*}[!htb]
  \centering
  \hspace*{-2.0cm}
  \includegraphics[width=0.4\linewidth]{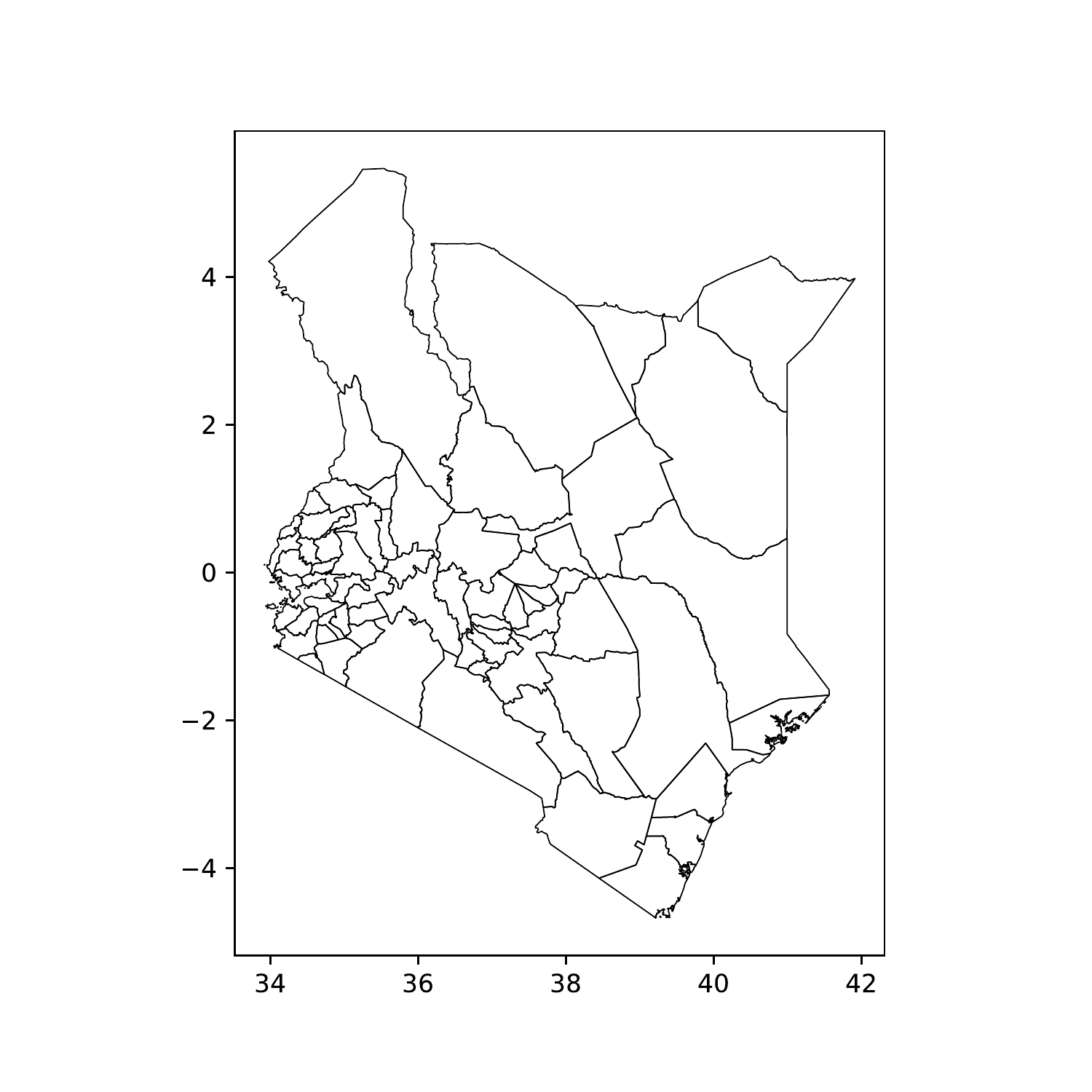}
  \hspace*{-2.0cm}
  \includegraphics[width=0.4\linewidth]{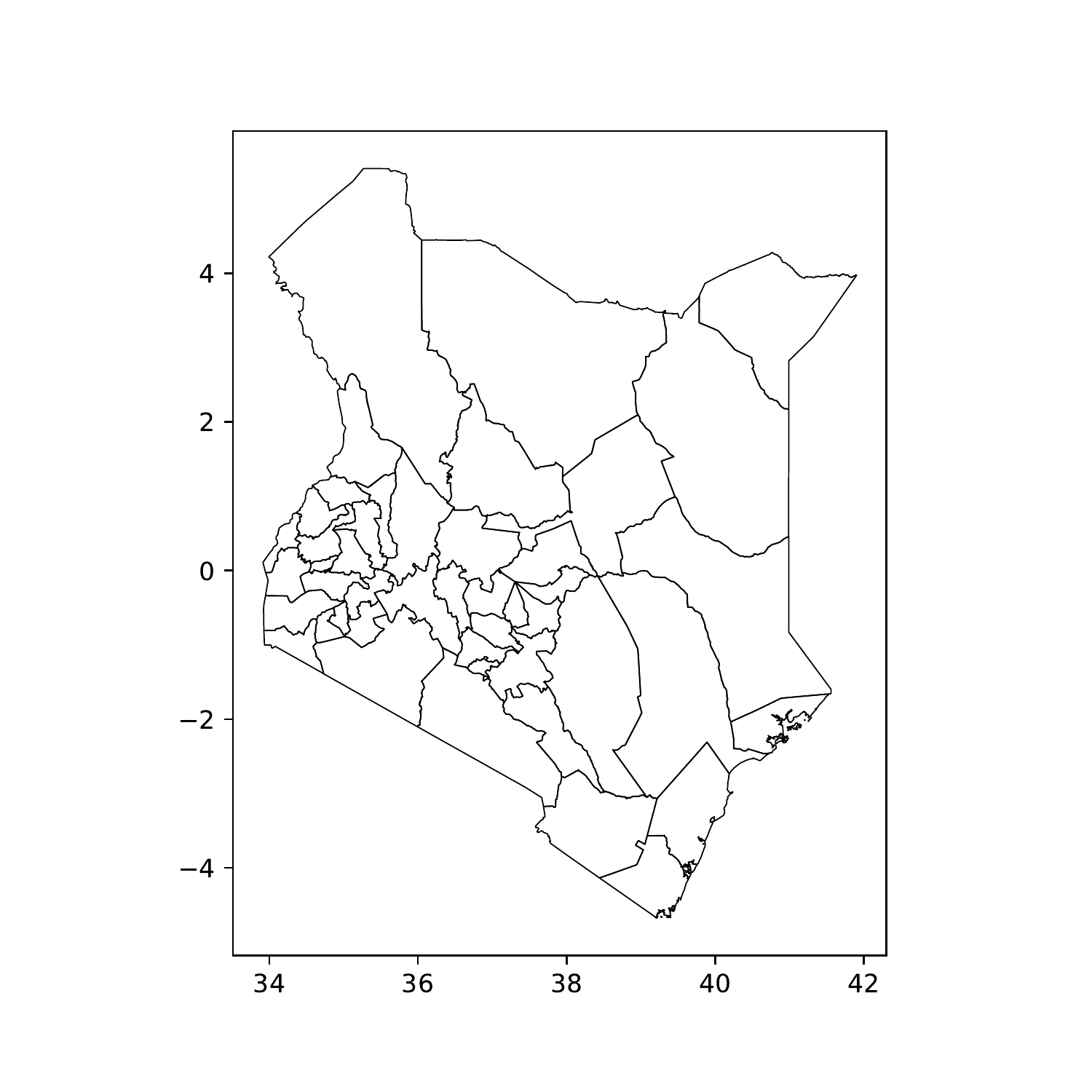}
  \hspace*{-2.0cm}
   \includegraphics[width=0.4\linewidth]{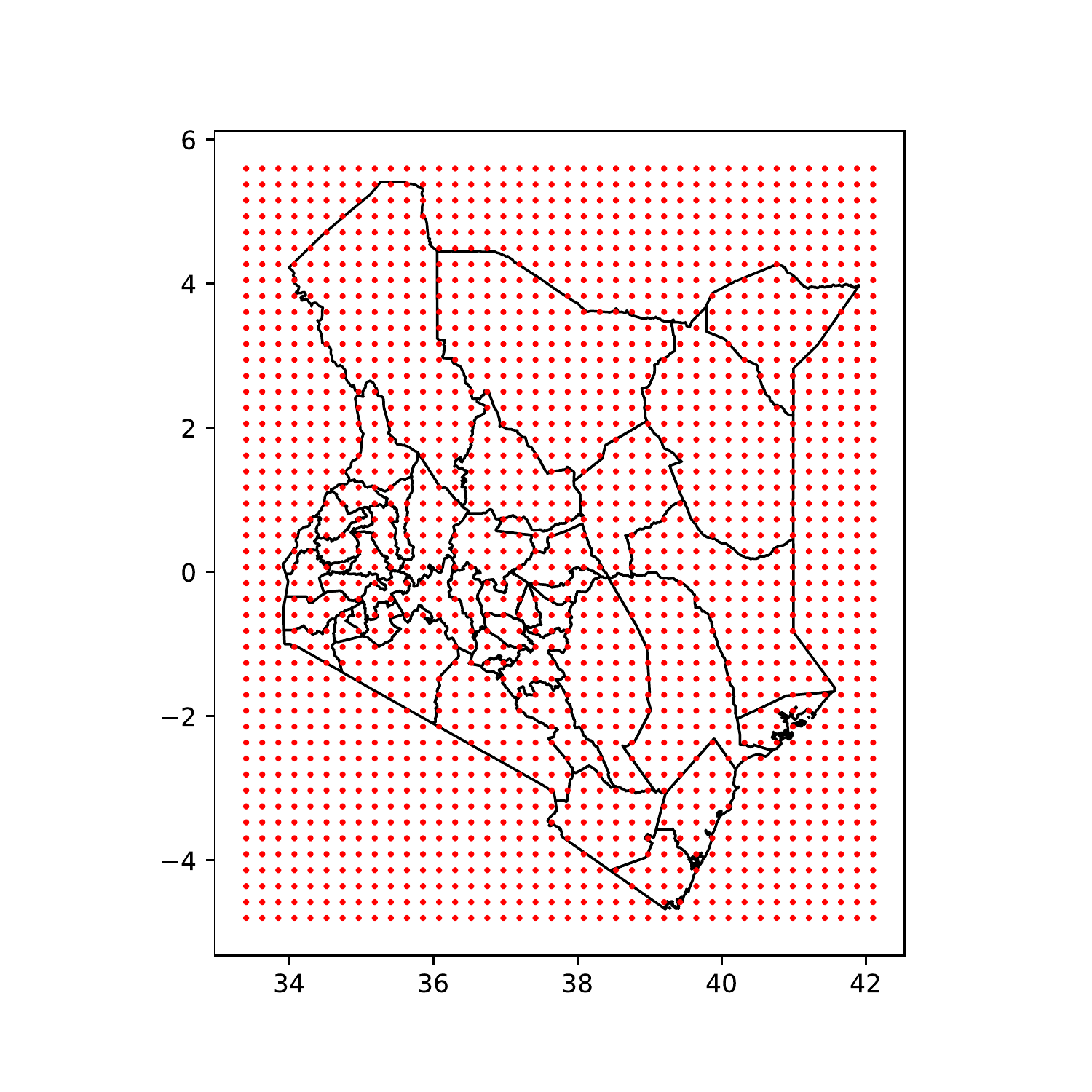}
  \caption{Map of Kenya with district borders before 2010 (left), district borders from 2010 (middle), and computational grid used (right).}\label{fig:old_new}
\end{figure*}

Malaria is one of the major causes of mortality in sub-Saharan Africa, with a disproportionate burden on young children. In Kenya, a country with a long history of malaria control, approximately 75\% of the population was still at risk in 2022 \citep{kenya2022}. As malaria control programs continue to create novel control strategies, district-level disease mapping remains a fundamental surveillance tool for analysing the present and historical distribution of the disease in both space and time. However, disease tracking becomes more difficult in the situation of crises. For example, political factors have historically driven decentralisation across developing countries often leading to changes in administrative boundaries. Many of countries have increased their number of sub-national administrative units, including more than twenty countries in sub-Saharan Africa \citep{hassan2016state}. Some countries have experienced multiple changes of boundaries, such as Kenya. Methodologically, district-level disease mapping in Kenya can be challenging because administrative boundaries changed in 2010: while the old system consisted of 8 provinces and 69 districts (Figure~\ref{fig:old_new}, left), the new system contains 47 districts (Figure~\ref{fig:old_new}, middle) which do not coincide with the old boundaries. This change is hard to tackle with standard disease mapping tools. 

Hierarchical Bayesian models are the state-of-the-art approach for disease mapping \citep{macnab2022bayesian, kang2016making, wakefield2000bayesian} since they are able to capture structure in the data, as well as to characterise uncertainty. Modern literature builds on a series of foundational works by \cite{besag1974spatial, clayton1992bayesian, bernardinelli1992empirical, bernadinelli1997disease, clayton1993spatial} who instigated a paradigm shift from the frequentist to the Bayesian approach in disease mapping. Bayesian disease mapping since the 1980s and 1990s has Bayesian hierarchical models as its foundation, with no shortage of examples in malaria mapping \citep{gemperli2006mapping, gosoniu2006bayesian,  hay2009world, reid2010mapping, bhatt2015effect, bhatt2017improved, snow2017prevalence, weiss2019mapping}. Markov chain Monte Carlo (MCMC) simulation methods, especially as implemented in probabilistic programming languages like BUGS and Stan, are a common approach to estimation, learning, and inference of unknown quantities and parameters. In recent years approximate inference algorithms, such as integrated nested Laplace approximation (INLA) have gained popularity \citep{martins2013bayesian}. While such tools provide convenient interfaces to a set of predefined (and well documented) models, they do not provide as much flexibility for custom model development as probabilistic programming languages. This limits their applicability to specific classes of applied research, including the problem of  aggregation and change-of-support which we study here. MCMC, on the contrary, is a general sampling technique for sampling from a possibly unnormalised target density $\pi(z)$, i.e.~the posterior \citep{robert1999monte, gelman1995bayesian}. 

MCMC (Markov Chain Monte Carlo) is a computational method that generates a sequence of samples from a target distribution by constructing a specific kind of Markov chain that is guaranteed to converge to the target distribution asymptotically. In practice, MCMC sampling is performed for a finite number of iterations $N$, where $N$ is sufficiently large for MCMC to converge. MCMC convergence and efficiency can be assessed via diagnostic tools, such as the $\hat{R}$ statistic and effective sample size (ESS) metrics, respectively \citep{vehtari2021rank}. The $\hat{R}$ statistic, also known as the potential scale reduction factor, is a measure used to assess the convergence of multiple MCMC chains by comparing the within-chain variance to the between-chain variance. The ESS statistic quantifies the amount of independent information obtained from the generated samples, reflecting the effective size of the sample in terms of capturing the true underlying distribution. The critically important propertie of MCMC is that in the asymptotic limit, it  guarantees  \textit{exact} samples from the target density. In health policy-related  applications where modelling informs decision-making this property makes MCMC preferable to non-exact methods (such as variational Bayes). However, MCMC scales poorly for problems involving correlation structures, such as Gaussian Processes (GPs). Additional issues inherent to MCMC are autocorrelation in the produced samples, meaning that chains must be simulated for a prohibitively long time in order to obtain reliable uncertainty estimates. It is highly desirable for disease mapping models to retain modelling flexibility and reliability of MCMC, and, at the same time, to improve inference speed and efficiency.

The main modelling tool for capturing spatial correlation in a disease mapping model is the Gaussian process (GP) prior, whose realisation over a finite number of points is a multivariate Normal ($\mathcal{MVN}$) distribution mean and covariance functions. The most common type of spatial data analysed in this context is  \textit{areal} data, i.e.~data obtained via aggregation of individual observations over spatial areas, such as administrative units. Statistical models describing areal data typically rely on the adjacency structure of areal units to account for spatial correlation. One drawback of this approach is that it disregards the continuous nature of underlying processes and potential heterogeneity within each region, especially large ones. Additionally, adjacency-based methods are very rigid with respect to the change-of-support problem, i.e.~when administrative boundaries change or when mapping needs to be done at a different administrative level. The spatial aggregation process has been proposed in the literature to address this issue: an observational model is designed using the integration of the GP over the corresponding region
\citep{tanaka2019spatially, yousefi2019multi, zhu2021aggregated, johnson2019spatially}. 

In this work, we present a novel, practical and easy to implement solution of the change-of-support problem relying on a methodology combining deep generative modelling and fully Bayesian inference. Our approach is twofold: 
\begin{itemize}
    \item We view the spatial process as continuous. Rather than performing modelling based on the adjacency structure, we model the latent GP process on a fine spatial scale over an artificial computational grid covering the domain of interest and obtain unit-level estimates via aggregation. This approach has been used before in order to better capture continuity than adjacency-based methods. 
    \item We extend the recently proposed PriorVAE \citep{semenova2022priorvae} method of encoding spatial priors with variational autoencoders to the change-of-support problem and malaria prevalence mapping in Kenya. Realisations of GP priors are generated on the fine spatial grid, and then aggregated to the level of administrative units. The aggregated values are encoded using the PriorVAE technique. The trained priors, termed aggVAE, are then used at the MCMC inference stage instead of combining the generation of GP priors and the aggregation step at each MCMC iteration. 
\end{itemize}
We show that MCMC using the aggVAE approximate prior  is faster and more efficient within an MCMC inference scheme than MCMC relying on the exact GP prior.

Our paper is structured as follows: in Section \ref{subsec:stats_models} we describe models from classical spatial statistics used to analyse areal data. In Section \ref{subsec:vaes} we introduce the field of deep generative modelling and in particular the VAE architecture. In Section \ref{subsec:prior_vaes} we summarise the PriorVAE method of encoding spatial priors. In Section \ref{sec:contribution} we propose the aggVAE method allowing to encode aggregated latent GP evaluations. Our application to malaria prevalence mapping in Kenya is presented in Section  \ref{sec:inference_kenya} and we conclude by discussing limitations and future work in Section \ref{sec:discussion}.

\section{Background}
\subsection{Spatial statistics models of areal data}
\label{subsec:stats_models}
Classical statistical models describing areal data typically rely on the adjacency structure of areal units to account for spatial correlation, that is, near by regions will likely be similar to each other. The prior on the spatial term in such models can be written as:
$$f \sim \mathcal{MVN}(0, Q^{-1}),$$ where $Q$ denotes the inverse covariance or precision matrix. Adjacency characterises the neighborhood structure allowing to calculate $Q$ based on the connectedness of the adjacent graph. These methods take advantage of the tendency for neighboring areas to possess similar features. \cite{besag1974spatial} first proposed the Conditional Auto-Regressive (CAR) with 
$$Q = \tau (I - \alpha A),$$ where $\tau$ denotes the marginal precision, $A$ is the adjacency matrix and $\alpha$ is a parameter capturing the amount of spatial dependence. Variations of this model were later proposed and include intrinsic CAR (iCAR): \citep{besag1991bayesian} 
$$Q = \tau (D-A),$$ 
where $D$ is the diagonal matrix consisting of the total number of neighbours for each area, proper CAR (pCAR)  \citep{cressie2015statistics} with 
$$Q = \tau (D-\alpha A),$$ Leroux CAR (LCAR) \citep{leroux2000estimation} with 
$$Q = \tau(\alpha(D-A) + (1-\alpha) I),$$ Besag-York-Mollié (BYM) model \citep{besag1991bayesian}:  
$$Q=\frac{1}{\tau_s}(D - A) + \frac{1}{\tau_\text{iid}}I,$$ BYM2 \citep{riebler2016intuitive}.

\subsection{Variational autoencoders (VAEs)}
\label{subsec:vaes}
A Variational Autoencoder (VAE) is a type of generative model that uses deep learning techniques to generate new data samples $\hat{y} \in \mathcal{Y} \subset \mathbb{R}^n$ that resemble the original training data $y \in \mathcal{Y}$. It consists of two parts: an encoder $E_\phi(.)$ that maps input data to a lower-dimensional representation (latent space) $\mathcal{Z} \subset \mathbb{R}^d, \quad d<n$, and a decoder $D_\psi(.)$ that maps the latent representation $z \in \mathcal{Z}$ back to the original data space. The encoder and decoder are trained together to minimise a reconstruction loss, which measures the difference between the original data and its reconstructed version. Additionally, a constraint is imposed on the latent representation to follow a prior distribution $q(z|y)$, such as a Gaussian distribution, allowing the model to generate new, unseen data by sampling from the prior and passing it through the decoder. Following \cite{kingma2013auto}, the optimal parameters for the encoder and decoder are found by maximising the evidence lower bound, or, equivalently, minimising the loss: 
$$\mathcal{L}_\text{VAE} = \mathbb{E}_{q(z|y)} \left[-\log p (y | z)\right] + KL \left[ q(z|y)|| p(z) \right].$$ The prior of the latent space and the variational distribution are typically chosen to have Gaussian forms: $p(z)=\mathcal{N}(0, I_d), \quad q(z|y) = \mathcal{N}(\mu_z, \sigma_z^2 I_d).$

\subsection{Encoding spatial priors with VAEs}
\label{subsec:prior_vaes}

\begin{figure*}[!t]
  \begin{center}
   \hspace*{-2.0cm}
  \includegraphics[width=0.4\linewidth]{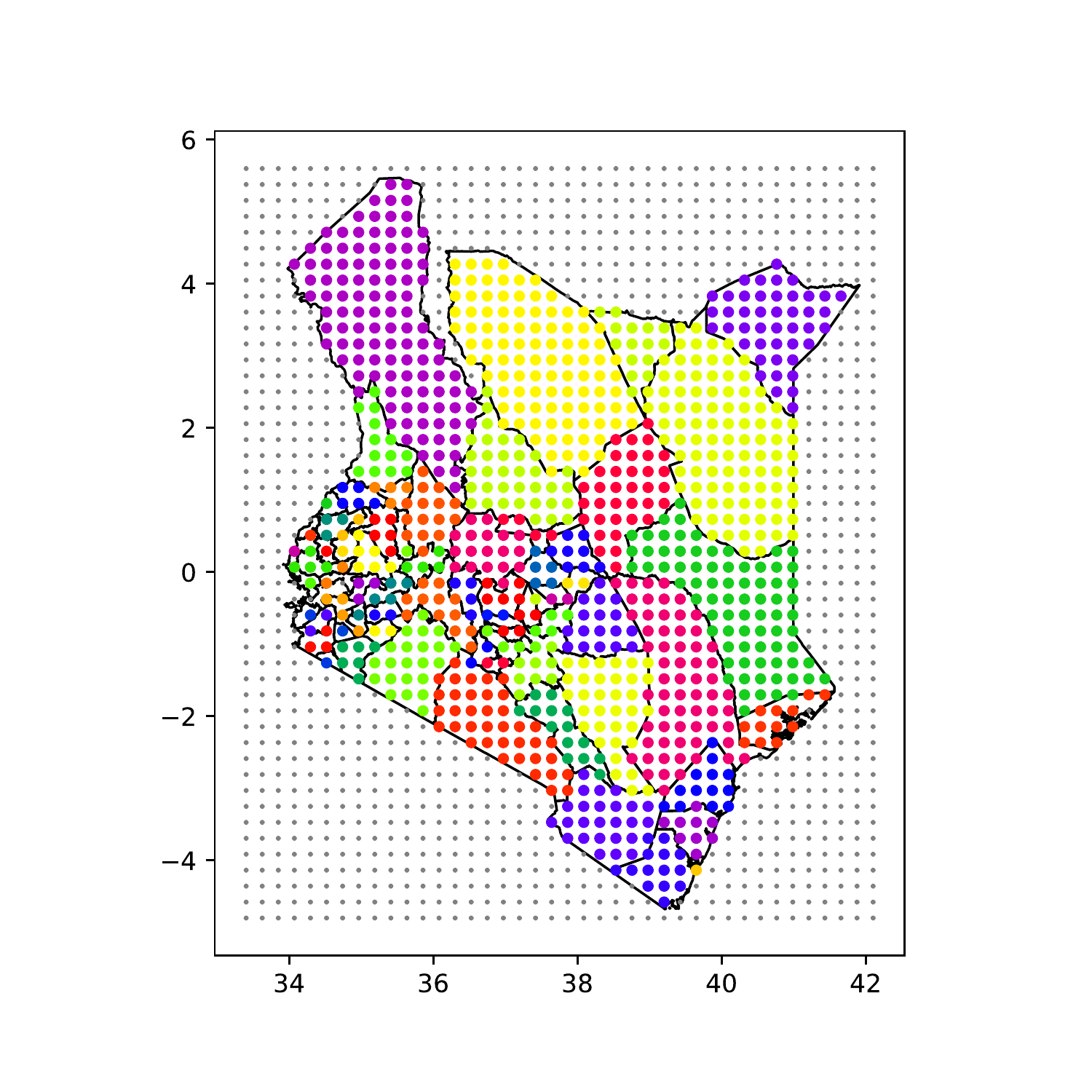}
   \hspace*{-2.0cm}
  \includegraphics[width=0.4\linewidth]{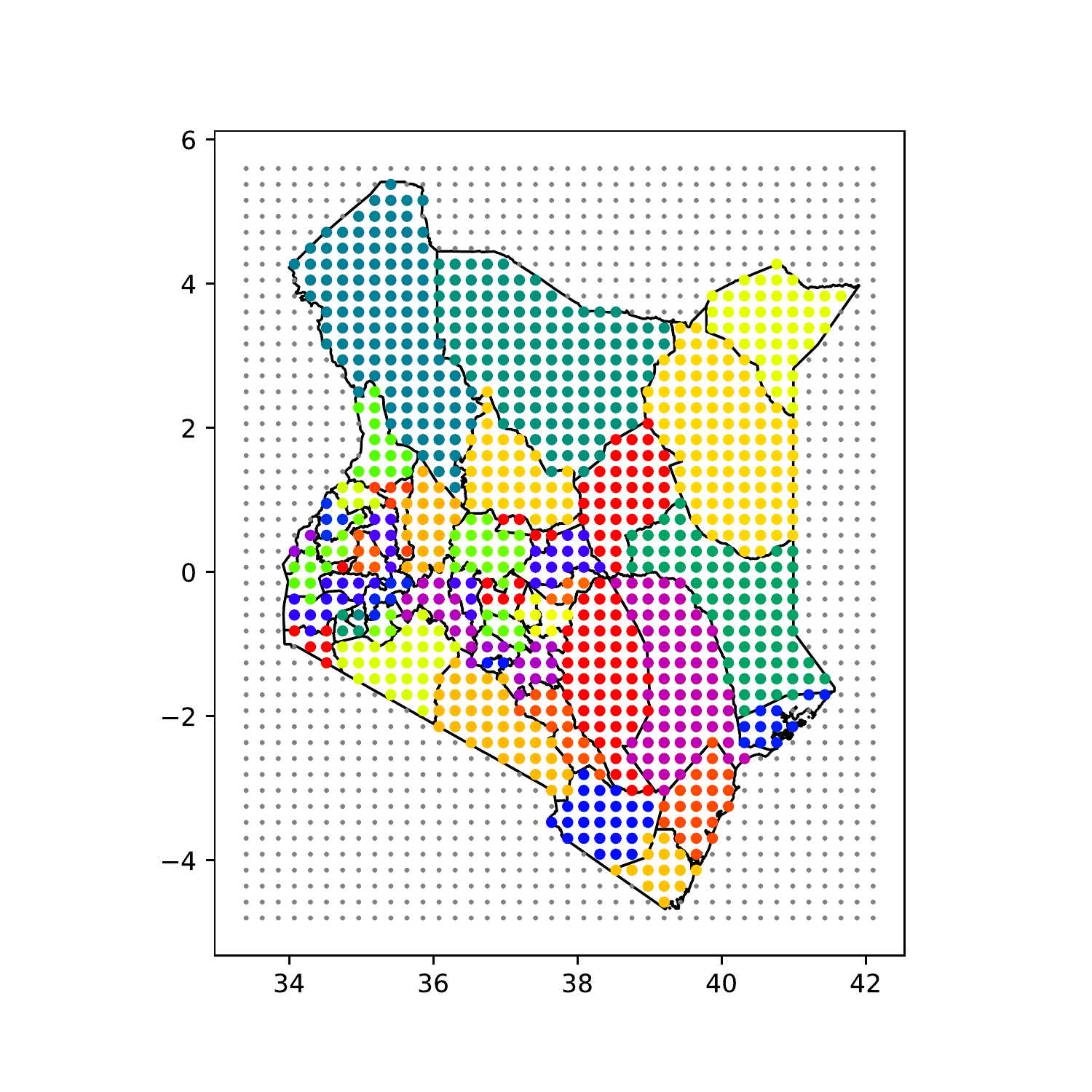}
  \end{center}
  \caption{Attribution of grid points over polygons (districts) before 2010 (left), and after 2010 (right). Grey points correspond to points falling outside of the country's borders, and points falling within the same polygon are represented with the same color.} \label{fig:grid_poly}
\end{figure*}
$\pi$VAE \citep{mishra2020pi} and PriorVAE \citep{semenova2022priorvae} are two related VAE-based methods that can respectively encode continuous stochastic processes and their finite realisations. They utilize a trained decoder to approximate computationally complex GPs and $\mathcal{MVN}$s for Bayesian inference with MCMC, preserving the rigour of MCMC while ensuring scalability through the simplicity of the VAE's latent space. The key difference between the two methods is the type of prior they encode and the method of encoding: $\pi$VAE uses a low-dimensional embedding of function classes via a combination of a trainable feature mapping and a generative model, while PriorVAE directly encodes fixed, finite-dimensional GP realisations. In this work we will follow the PriorVAE  approach. The method has been proposed as a scalable solution to the small area estimation (SAE) problem in spatial statistics as it encodes realisations of priors presented in Section \ref{subsec:stats_models}. A characteristic property of PriorVAE is that it needs to be trained on a predefined spatial structure. On one hand, this is a disadvantage compared to $\pi$VAE since PriorVAE is unable to make predictions on off-grid locations. On the other hand, in the settings when the spatial structure is known in advance, PriorVAE is preferred due to its simpler computational setup, as only the prior-encoding VAE needs to be trained, without the need for learning of the feature map. The inference workflow using the PriorVAE method can be described as follows:
\begin{itemize}
    \item Fix the spatial structure of interest $\{x_1, \ldots, x_n\}$ -  a set of administrative units, or an artificial computational grid.
    \item Draw evaluations of a GP prior $\mathcal{GP}(.)$ over the spatial structure and use the vector of realisations $$f_\text{GP}=\left(f(x_1), \ldots, f(x_n)\right)^T$$ as data for a VAE to encode.
    \item Train a VAE on the generated data to obtain the decoder $D_\psi(.)$.
    \item Perform Bayesian inference of the overarching model using MCMC, where $f_\text{GP}$ is approximated  by the trained decoder $D_\psi(.):$ $$f_\text{GP} \approx \hat{f}_\text{GP} = f_\text{PriorVAE} = D_\psi(z_d), \quad z_d \sim \mathcal{N}(0, I_d).$$
\end{itemize}


\section{Encoding aggregates of the Gaussian Process prior: aggVAE}
\label{sec:contribution}
\subsection{GP evaluations over a fine scale grid}
We view the underlying process as continuous and approximate it by evaluating the GP on a fine spatial grid $G = \{g_1, ... g_n\}$ (Figure~\ref{fig:old_new} (right)) 
covering the domain of interest. The grid is regular and has been chosen to ensure that at least one point of the grid $g_j$ lies within each administrative unit $p_i$. GP prior realisations $f(.)$ are drawn over the artificial grid $G$ as a multivariate normal distribution with a covariance matrix following the RBF\footnote{Any kernel can be used. We use RBF only as an example.} kernel:
\begin{equation*}
    f = \begin{pmatrix}
           f_{1} \\
           \vdots \\
           f_{n}
         \end{pmatrix} \sim \mathcal{MVN}(0, \Sigma), \quad \Sigma_{jk} = \sigma^2 \exp \left( - \frac{d^2_{jk}}{2 l^2} \right)
\end{equation*} 
where $f_j = f(g_j),$ $d_{jk}=||g_j-g_k||$ and $\sigma^2, l$ are hyperparameters of the Gaussian Process. For the hyperparameters we used $l \sim \text{InvGamma}(3,3)$ and $\sigma \sim \mathcal{N}^+(0.05)$ priors.
\subsection{Computing GP aggregates over polygons}
As the next step, we aggregate GP evaluations to the district level. Each district is viewed as a polygon $p_i, i=1,...,K$, and the computation takes the form
\begin{equation}
    \label{eq:sum_polyg}
    f_\text{aggGP}^{p_i}=\int_{p_i} f(s)ds \approx c \sum_{g_j \in p_i} f_j = c \bar{f}_\text{aggGP}^{p_i}.
\end{equation}
Here $\bar{f}_\text{aggGP}^{p_i}=\sum_{g_j \in p_i} f_j$ and we have used the midpoint quadrature rule. The constant $c= \Delta x \Delta y$ with $\Delta x$ and $\Delta y$ being step sizes of the grid along the $x$ and $y$ axes, respectively. We can, therefore, construct a vector where each entry represents a spatial random effect at a district $p_i$:   
\begin{equation}
    \label{eq:f_gp_agg}
    f_\text{aggGP} = \begin{pmatrix}
           f_\text{aggGP}^{p_1} \\
           \vdots \\
           f_\text{aggGP}^{p_K} 
         \end{pmatrix} \in \mathbb{R}^K.
\end{equation}
In practice, we implement \ref{eq:sum_polyg} via matrix multiplication. For this, we precompute matrix $M$ consisting of $K$ rows and $n$ columns with binary entries $M_{ji},$ indicating whether point $j$ lies within polygon $i$ (see Figures \ref{fig:grid_poly}):
$$M_{ji}=I_{\{g_j \in p_i\}}, \quad j=1,...,K, \quad i=1,...,n.$$
Hence, $M$ serves as a lookup table, and if $f$ is a vector of GP draws over the grid, the product $M f$ gives the vector of sums
\begin{equation}
    \label{eq:f_gp_agg}
    \bar{f}_\text{aggGP} = Mf = \begin{pmatrix}
           \bar{f}_\text{aggGP}^{p_1} \\
           \vdots \\
           \bar{f}_\text{aggGP}^{p_K}. 
         \end{pmatrix} 
\end{equation}
This procedure can be performed both with respect to the old and new boundaries to obtain vectors $f_\text{aggGP}^\text{old}$ and $f_\text{aggGP}^\text{new}$, respectively, using $M^\text{old}$ and $M^\text{new}$ precomputed matrices. 


\subsection{Encoding GP aggregates}
In order to tackle the change-of-support problem, we encode $\bar{f}_\text{aggGP}^\text{old}$ and $\bar{f}_\text{aggGP}^\text{new}$ jointly. We construct a vector of dimension $K_1 + K_2$ of the form
\begin{equation*}
    \bar{f}_\text{aggGP}^\text{joint} = \begin{pmatrix}
    \bar{f}_\text{aggGP}^{p^\text{old}_1} \\
    \hdots \\
    \bar{f}_\text{aggGP}^{p^\text{old}_{K_1}}\\
    ---- \\
    \bar{f}_\text{aggGP}^{p^\text{new}_1} \\
    \bar{f}_\text{aggGP}^{p^\text{new}_{K_2}}\\
    \end{pmatrix}
    = \begin{pmatrix}
    M^\text{old} f\\
    M^\text{new} f
    \end{pmatrix}  \in \mathbb{R}^{K_1+K_2}.
\end{equation*}
and apply the PriorVAE method to $\bar{f}_\text{aggGP}^\text{joint}$, i.e.
we encode GP aggregates jointly for old and new boundaries with a VAE using a lower-dimensional representation with independent standard Gaussian components $z_1, ..., z_d,  \quad d < K_1 + K_2, \quad z_i\sim N(0,1)$. We denote the new prior of the area-level spatial effect as $f_\text{aggVAE}.$ This one-step prior can be used at the inference stage instead of the two step procedure where first evaluation $f_1, \dots, f_n$ need to be drawn and then aggregated to obtain $f_\text{aggGP}$. We summarise the encoding and MCMC inference procedure using aggVAE in Algorithm \ref{alg:aggVAE}.
\begin{algorithm}
\caption{Inference procedure using aggVAE}\label{alg:aggVAE}
\begin{itemize}
\item Fix spatial structure of areal units as a collection of polygons $P=\{p_1, \dots, p_K\}$.
\item Create an aritificial computational grid of sufficient granularity $G=\{g_1, \dots, g_n\}$.
\item Precompute the matrix of indicators $M, \quad M_{ji}=I_{\{g_j \subset p_i\}}$.
\item Draw GP evaluations over $G$ using a selected kernel $k(., .)$: $f=(f_1, \dots f_n)^T$.
\item Compute GP aggregates at the level of $P: f_\text{aggGP}=cMf$
\item Train PriorVAE on $f_\text{aggGP}$ (or $\bar{f}_\text{aggGP}$) draws to obtain $f_\text{aggVAE}$ priors
\item Use $f_\text{aggVAE}$ at inference within MCMC.
\end{itemize}
\end{algorithm}

\section{Inference using aggVAE: mapping malaria prevalence in Kenya}
\label{sec:inference_kenya}
Malaria prevalence is routinely mapped using disease surveillance data which was collected, for example, via the DHS programme. A number of survey clusters (households) are selected and individuals within a cluster get tested for the presence or absence of malaria parasite. Malaria prevalence can then be modelled as the probability of a positive test among all tests. In this work we use results of the survey conducted in 2015. In 2010 administrative boundaries in Kenya changed (\ref{fig:old_new}). We treat the 2015 data as static, i.e.~we assume that the same data was collected once before 2010 and once after 2010 by overlaying it with old and new boundaries. District-specific malaria prevalence $\theta_i, i \in 1, \dots K$ is inferred using the Binomial distribution
\begin{align}
\label{eq:inference_gp}
    \begin{cases}
          n^\text{pos}_i &\sim \text{Bin}(n^\text{tests}_i, \theta_i),\\
          \text{logit}(\theta_i) &= b_0 + f^{p_i}_\text{aggGP}.
    \end{cases}
\end{align}
where $n^\text{tests}_i$ and $n^\text{pos}_i$ are the number of total and positive RDT tests observed in district $i$, correspondingly. $f^{p_i}_\text{aggGP}$ in \ref{eq:inference_gp} is the two-step spatial prior requiring first the sampling of the GP draws and the subsequent aggregation step. To avoid this procedure, we approximate $f^{p_i}_\text{aggGP}$ with $f^{p_i}_\text{aggVAE}$ and use the following model for inference:
\begin{align}
\label{eq:inference_vae}
    \begin{cases}
          n^\text{pos}_i &\sim \text{Bin}(n^\text{tests}_i, \theta_i),\\
          \text{logit}(\theta_i) &= b_0 + sf^{p_i}_\text{aggVAE}.
    \end{cases}
\end{align}
The additional parameter $s$ is introduced here to account for us encoding $\bar{f}_\text{aggGP}$ rather than $f_\text{aggGP}$, as well as to prevent our VAEs from oversmoothing; the additional parameter can correct for that at the inference stage.

\begin{figure*}[!htb]
  \centering
  \includegraphics[width=0.8\textwidth, height=1.5cm]{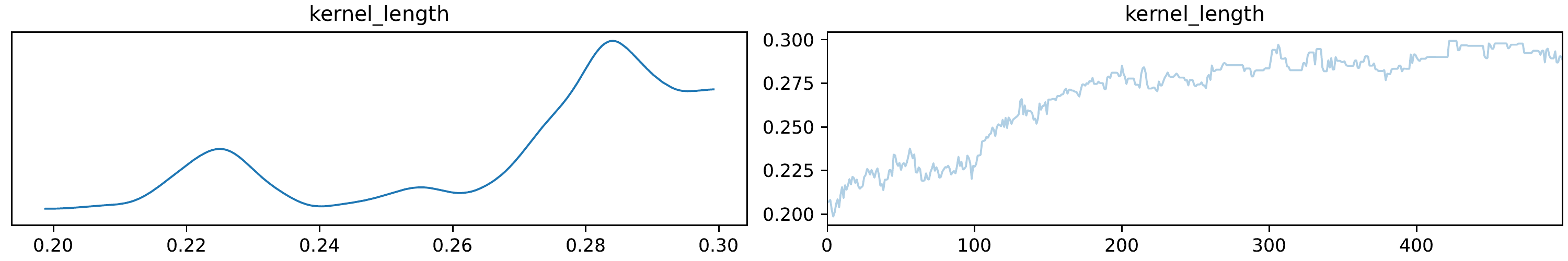}
  \caption{Posterior distribution (left) and traceplot (right) of the GP lengthscale parameter.}\label{fig:arviz_length}
\end{figure*}

\begin{figure*}[!htb]
  \centering
  \includegraphics[width=0.8\textwidth, height=1.5cm]{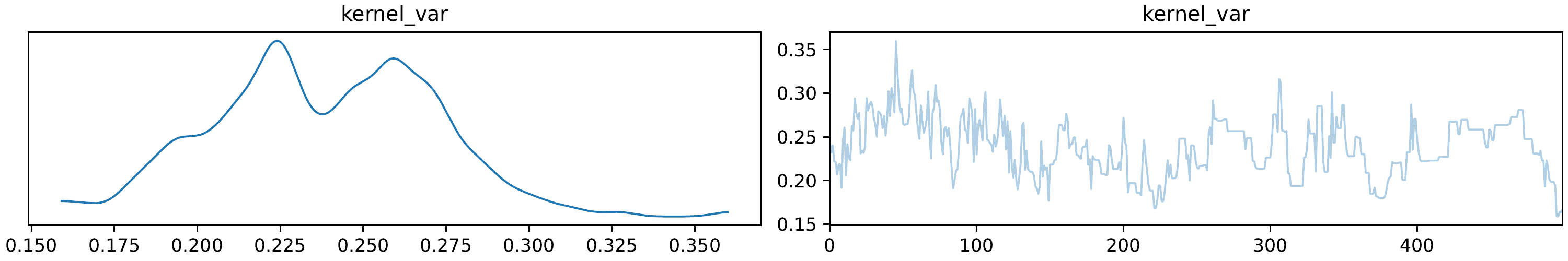}
  \caption{Posterior distribution (left) and traceplot (right) of the GP variance parameter.}\label{fig:arviz_var}
\end{figure*}

\begin{figure*}[!htb]
\vspace{-0.5cm}
  \centering
  \includegraphics[width=0.5\textwidth]{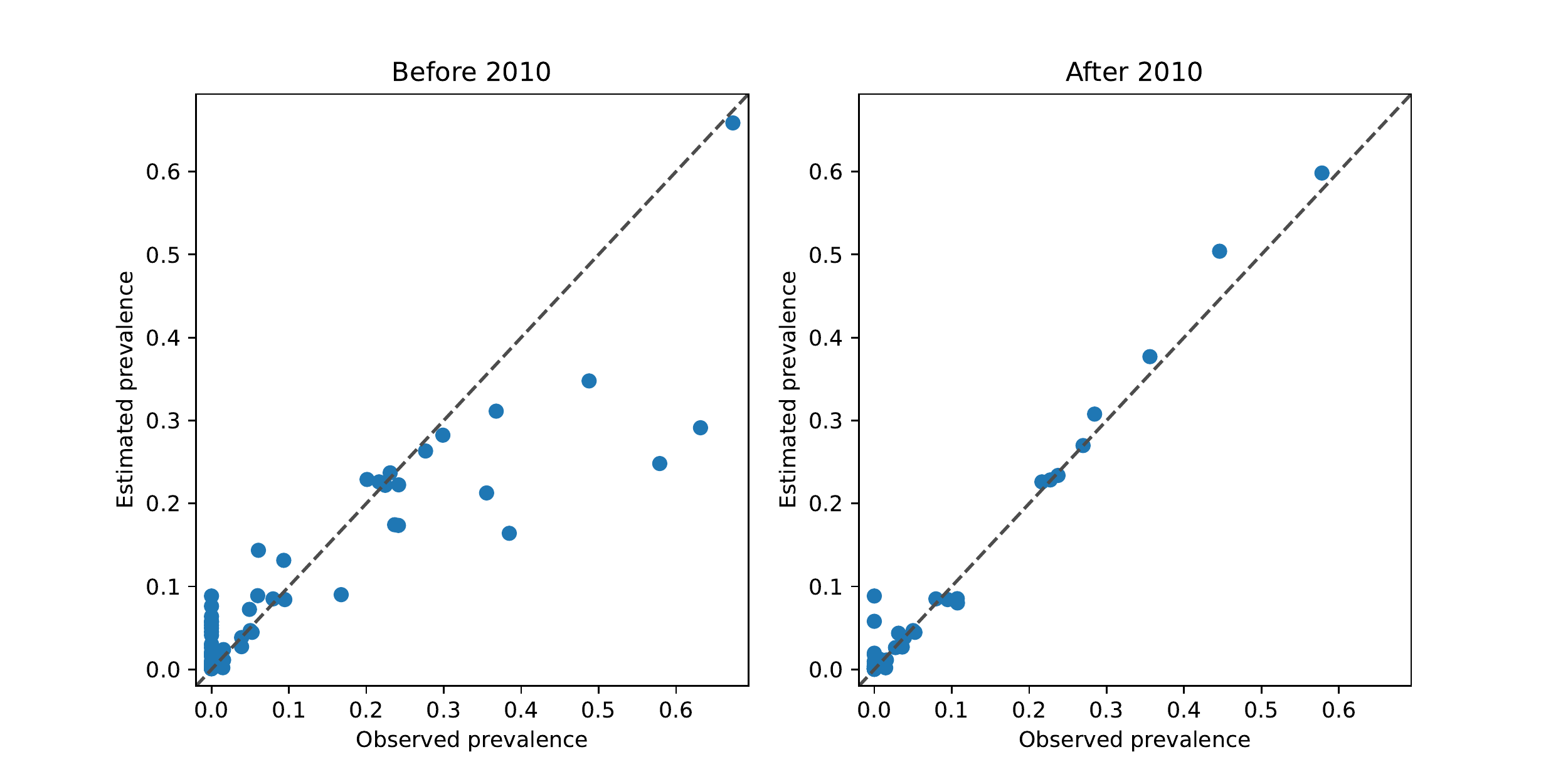}
  \caption{Observed and estimated prevalence produced by the aggGP model. The model has not converged after 1200 MCMC steps and particularly struggles with estimates over the old bounaries.}\label{fig:est_obs_gp}
\end{figure*}

\begin{figure*}[!htb]
  \centering
  \includegraphics[width=0.5\textwidth]{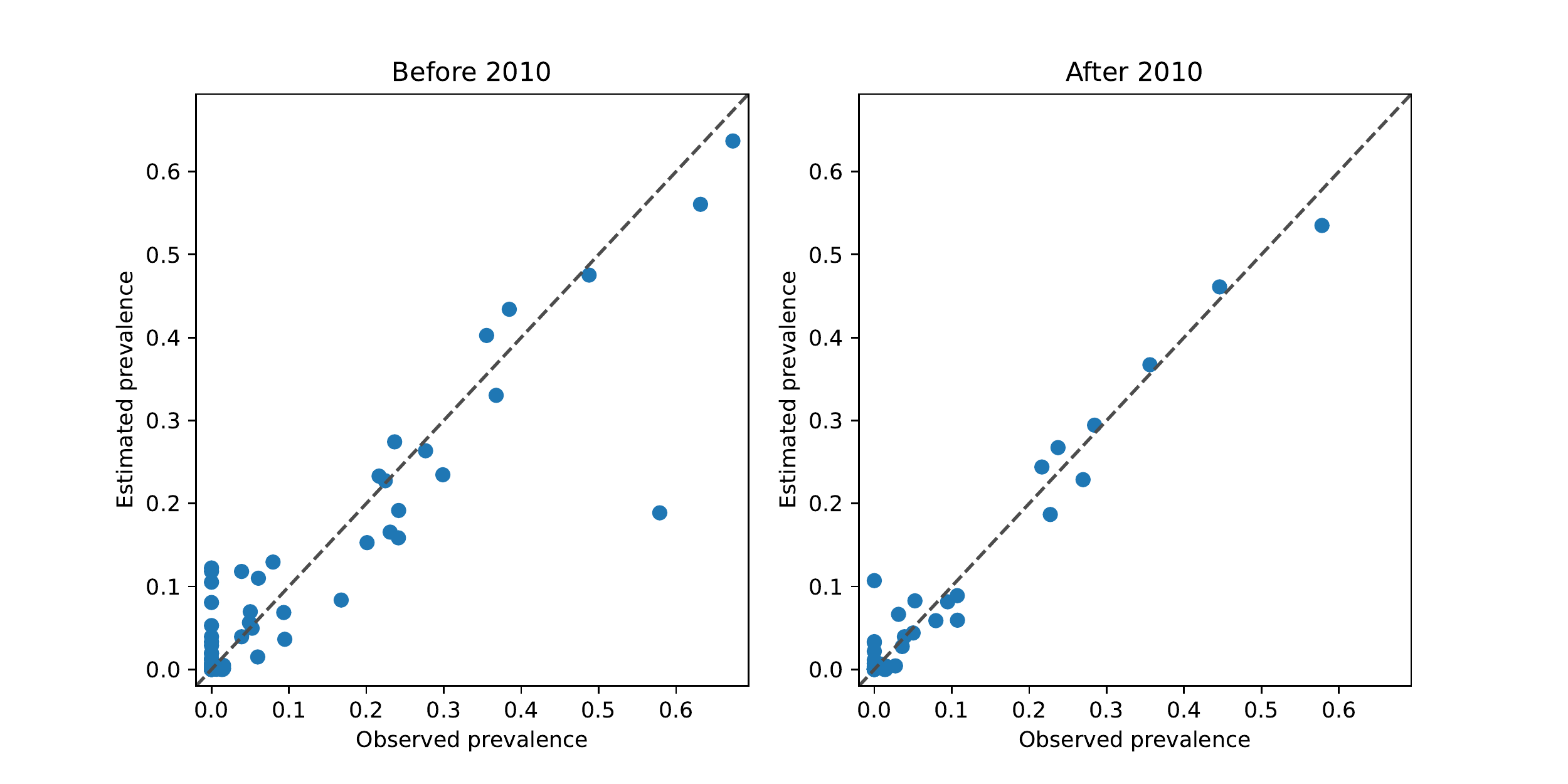}
  \caption{Observed and estimated prevalence produced by the aggVAE model.}\label{fig:est_obs_vae}
\end{figure*}

\begin{figure*}[!htb]
  \centering
  \hspace*{-2.0cm}
  \includegraphics[width=0.4\textwidth]{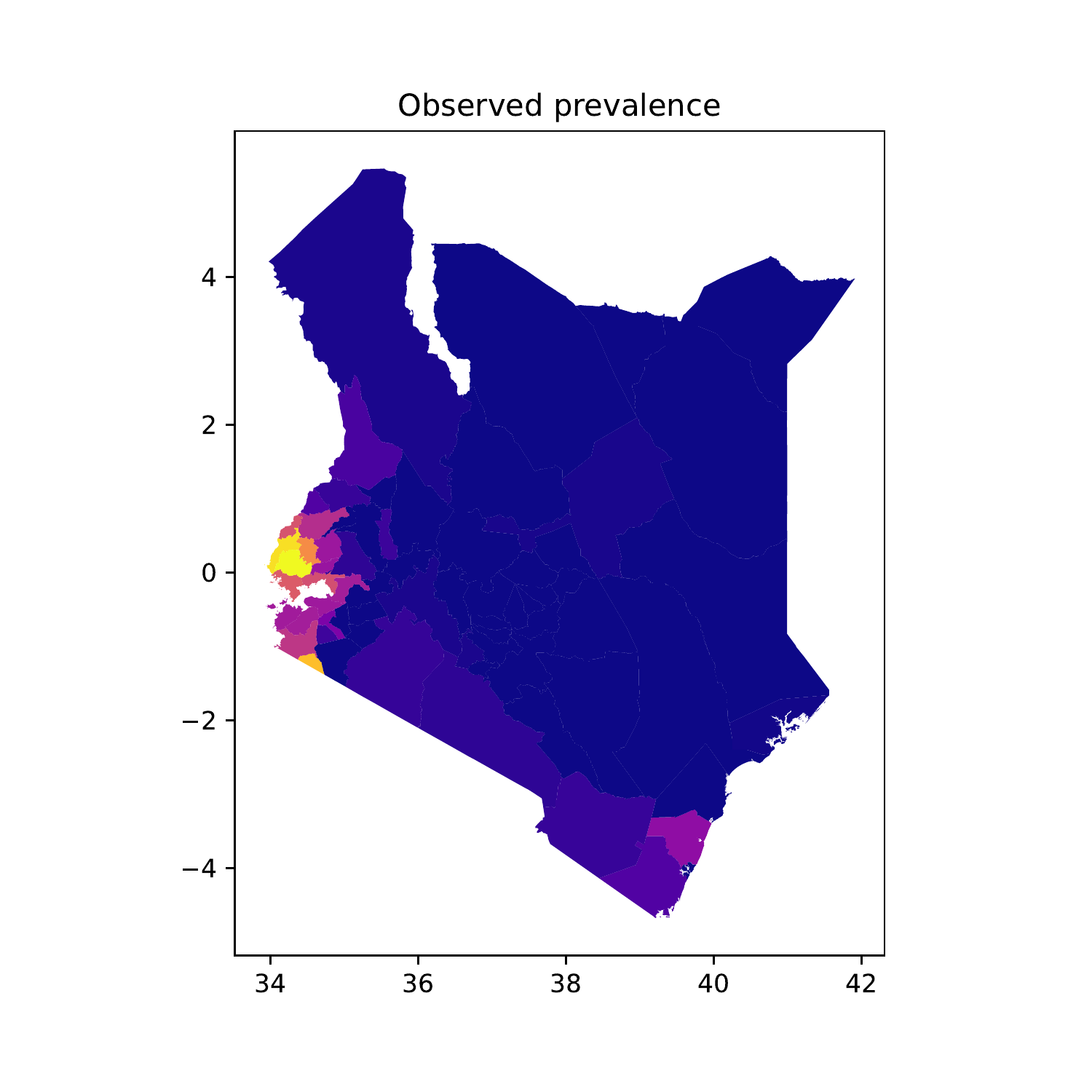}
  \hspace*{-2.0cm}
  \includegraphics[width=0.4\textwidth]{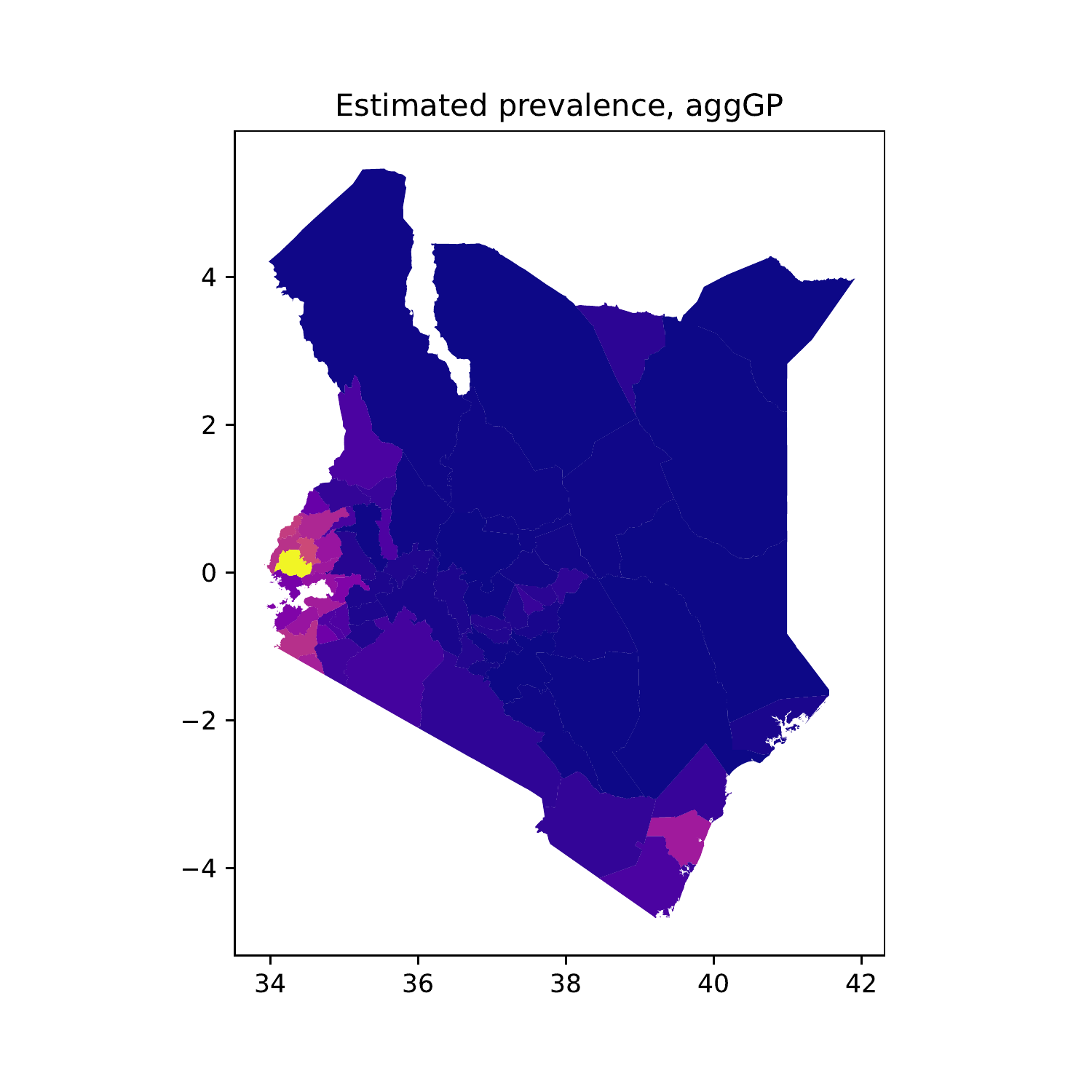}
  \hspace*{-2.0cm}
  \includegraphics[width=0.4\textwidth]{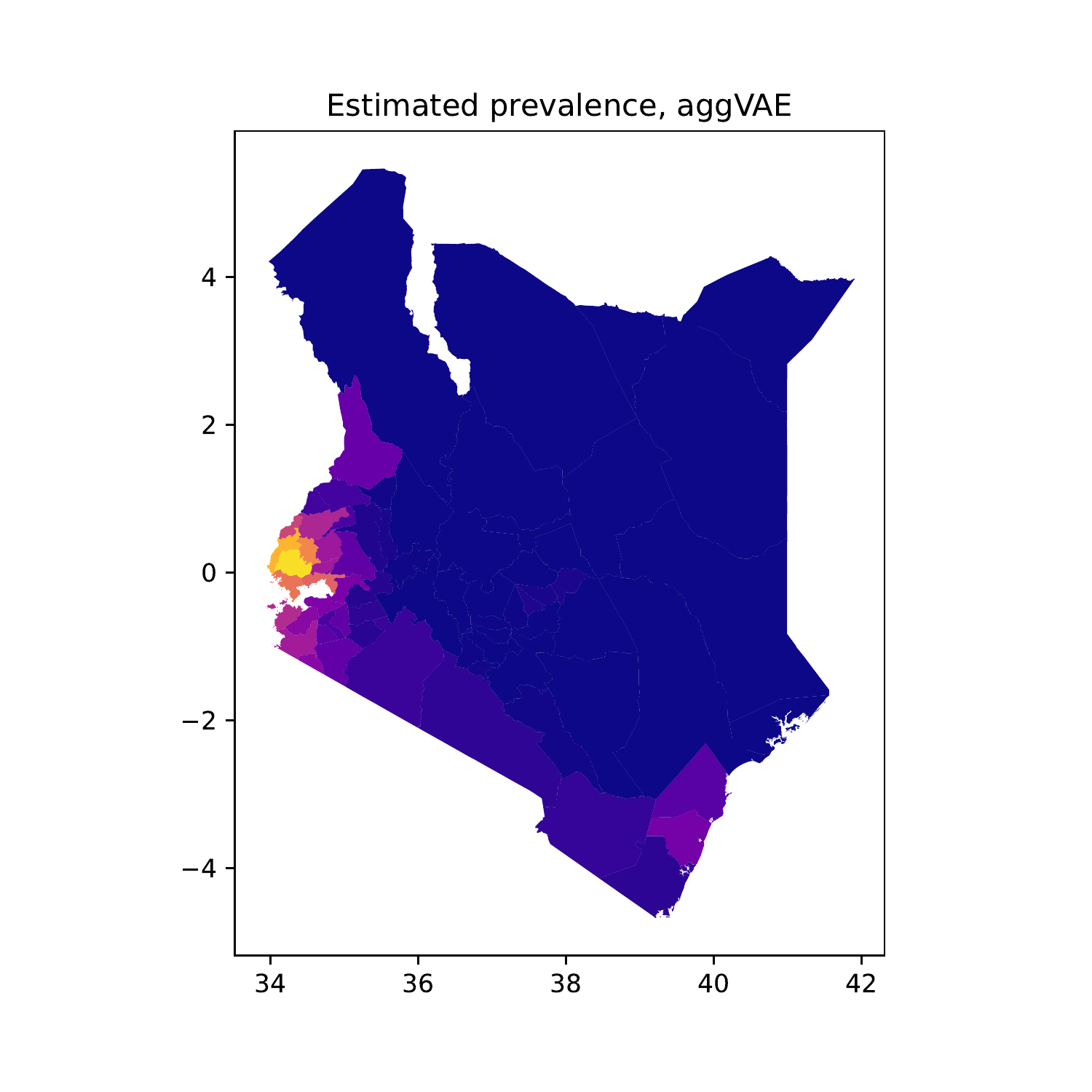}
   \hspace*{-1.2cm}
  \includegraphics[width=0.1\textwidth]{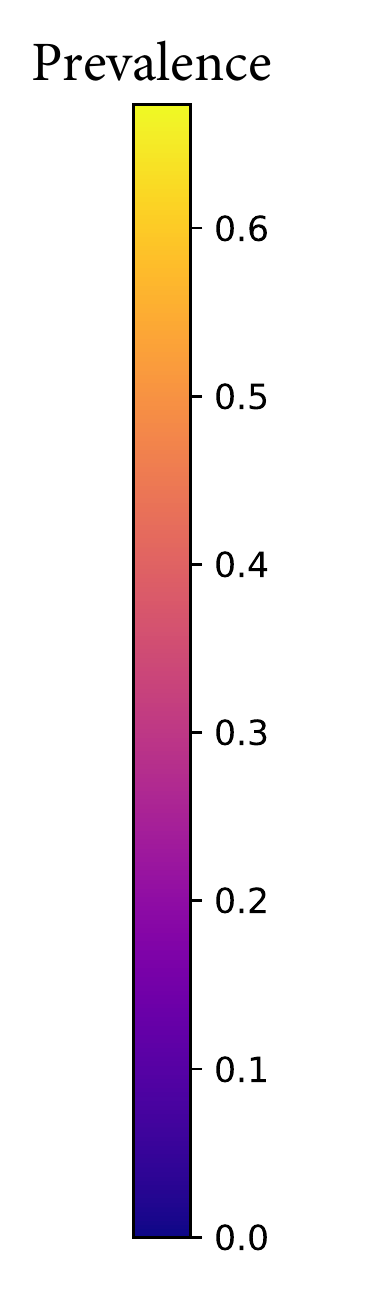}
  \caption{Map of malaria prevalence in Kenya based on district boundaries before 2010: (a) crude prevalence estimates, (b) estimates obtained by the aggGP model, and (c)  estimates obtained by the aggVAE model.}\label{fig:maps_before}
\end{figure*}

\begin{figure*}[!htb]
  \centering
  \hspace*{-2.0cm}
      \includegraphics[width=0.4\textwidth]{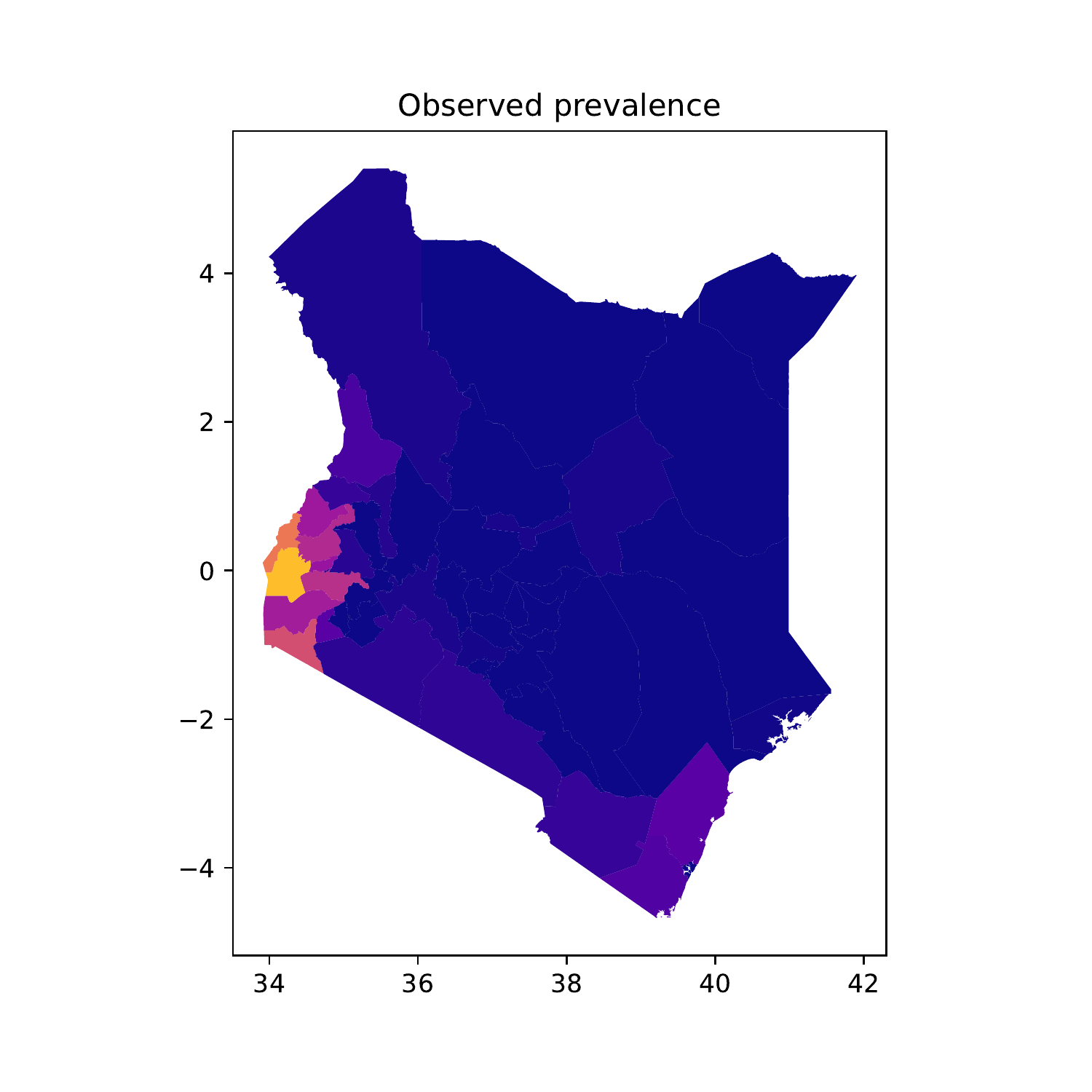}
   \hspace*{-2.0cm}
       \includegraphics[width=0.4\textwidth]{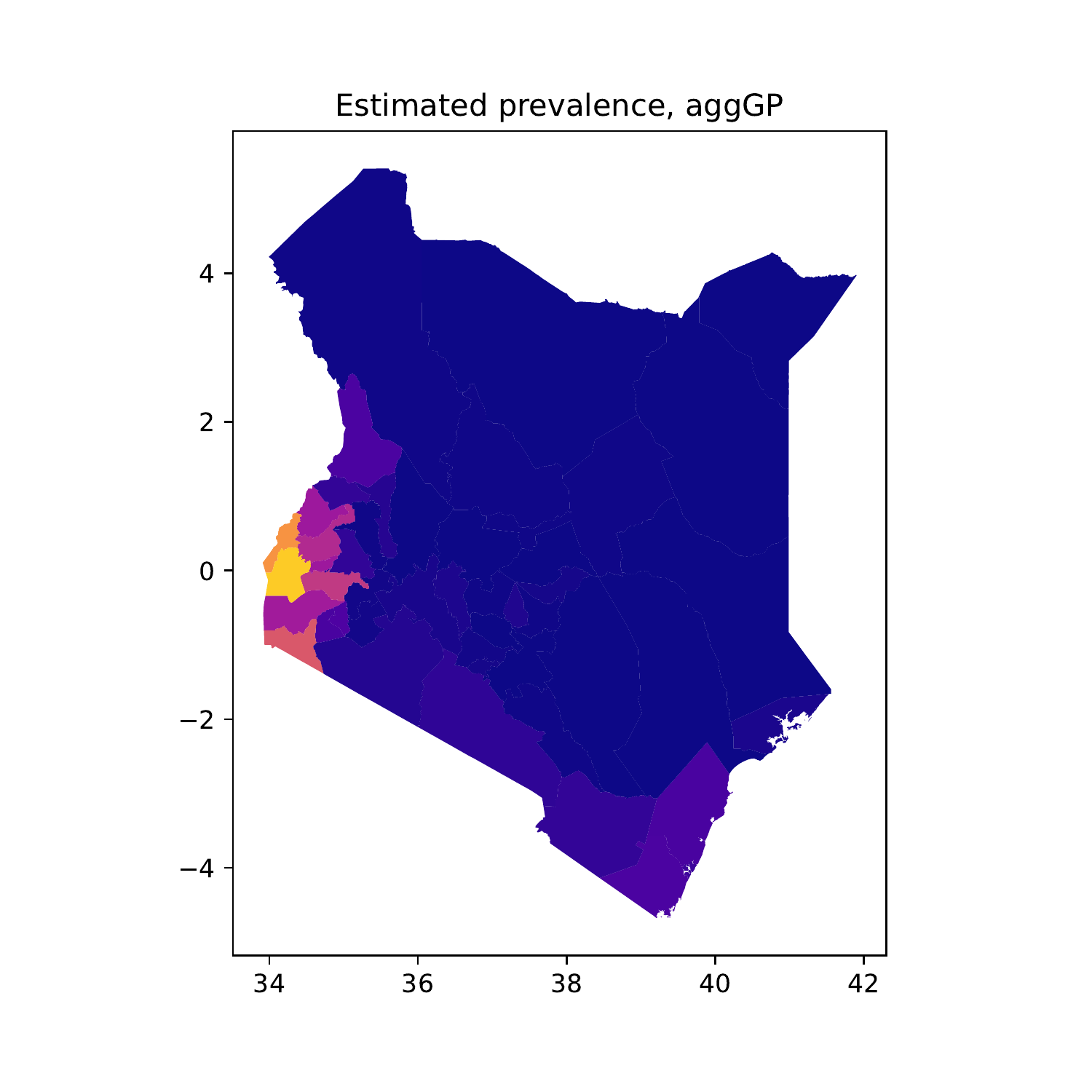}
    \hspace*{-2.0cm}
        \includegraphics[width=0.4\textwidth]{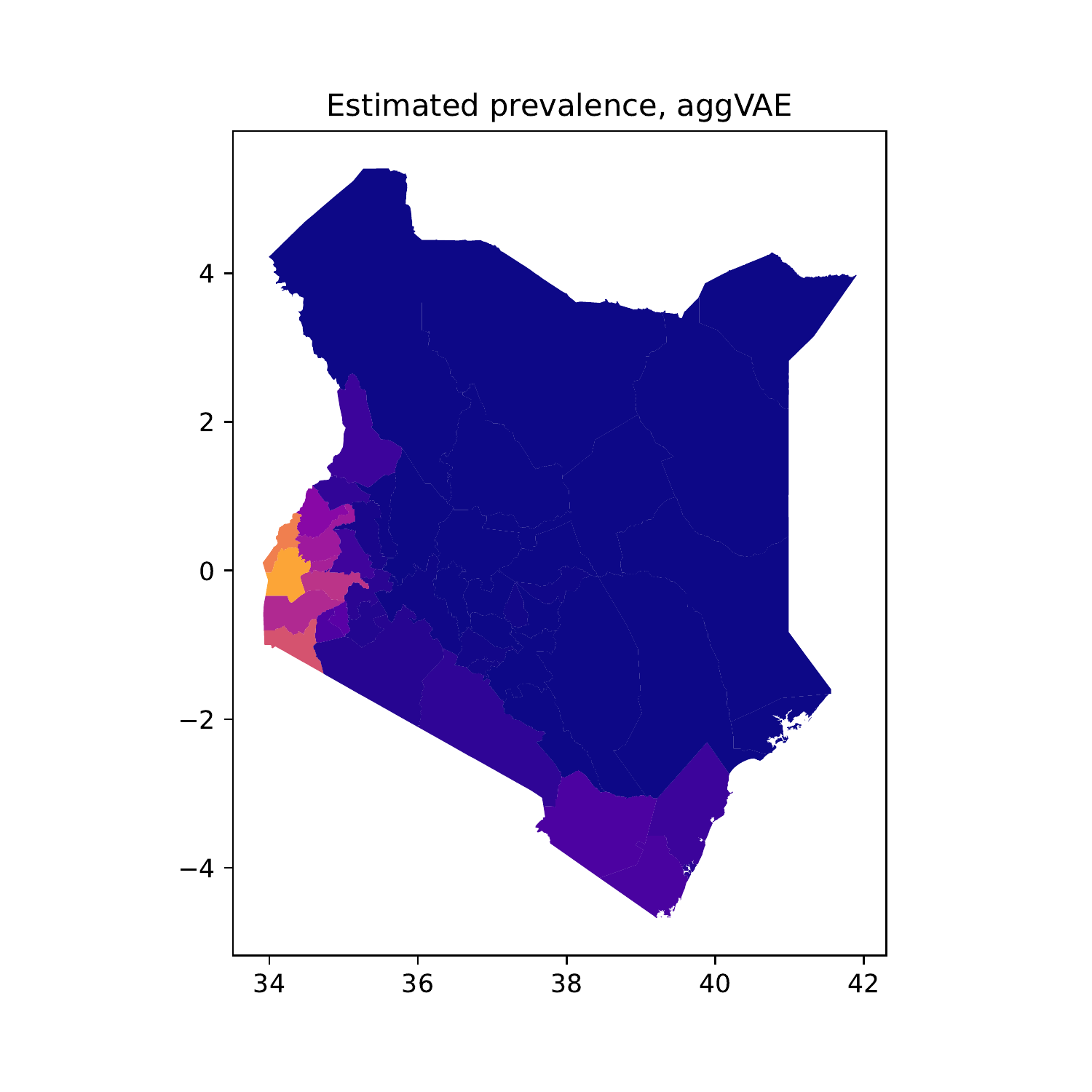}
    \hspace*{-1.2cm}
      \includegraphics[width=0.1\textwidth]{figures/colorbar.pdf}
  \caption{Map of malaria prevalence in Kenya based on district boundaries after 2010: (a) crude prevalence estimates, (b) estimates obtained by the aggGP model, and (c)  estimates obtained by the aggVAE model.}\label{fig:maps_after}
\end{figure*}

Our goal is to compare speed and efficiency in terms of effective sample sizes (ESS) of MCMC inference using models described in \ref{eq:inference_gp} and \ref{eq:inference_vae}. Both inference models were implemented using the Numpyro probabilistic programming language \citep{phan2019composable, bingham2019pyro} and the encoding of aggVAE was performed using the JAX library \citep{jax2018github}. We ran both MCMC inference models using 200 warm-up and 1000 posterior samples. Results of the comparison are presented in Table \ref{table:model-comparison}.  The model using aggGP prior ran $~$10K times longer, and after 14h has not fully converged. After 1200 total iterations it has only achieved ~$\hat R = 1.4$. Traceplots and posterior distributions for the GP lengthscale and variance parameter are presented on Figures \ref{fig:arviz_length} and \ref{fig:arviz_var}, correspondingly. When comparing spatial random effects (REs) corresponding to old and new boundaries, aggGP model particularly struggles with the old ones: maximum Gelman-Rubin statistic is $\hat{R}=1.10$ for REs over the old boundaries, and $\hat{R}=1.06$ for RFs over the new boundaries. Graphical comparison of the crude estimates, i.e. observed crude prevalence ($\theta_\text{crude} =\frac{n_i^\text{pos}}{n_i^\text{tests}}$) and estimates obtained by the aggGP and aggVAE models are presented on Figures \ref{fig:est_obs_gp} and \ref{fig:est_obs_vae}, correspondingly. Maps of crude prevalence estimates, estimates obtained by the aggGP and aggVAE models are presented on Figure \ref{fig:maps_before} for boundaries before 2010 and  on Figure \ref{fig:maps_after} for boundaries from 2010.

\begin{table}
    \centering
    \caption{Comparison of MCMC for models with $f_\text{aggGP}$ and $f_\text{aggVAE}$ spatial random effects (REs) using 200 warm-up and 1000 steps}\label{table:model-comparison}
    \begin{tabular}{lrr}
      \toprule 
      \bfseries  \shortstack{Model\\ (used prior)} & \bfseries  \shortstack{aggGP\\ RE} &  \bfseries \shortstack{aggVAE\\ RE}\\
      \midrule 
        \shortstack{Elapsed time} & 14h\tablefootnote{After this time aggGP model has not fully converged; e.g. the Gelamn-Rubin statistic for the lengthcsale parameter is ~$\hat R = 1.4$.} &  8s\\[1mm] 
  \shortstack{Average ESS of the REs} 
     & 129 &  231\\[1mm] 
  ESS per minute & 0.15 &  1732\\ \hline
  \shortstack{Maximum $\hat{R}$ of REs,\\boundaries before 2010} & 1.10 & 1.01\\[2mm] 
  \shortstack{Maximum $\hat{R}$ of REs,\\boundaries from 2010} & 1.07 & 1.01\\ \\ 
  \shortstack{Average ESS of the REs,\\boundaries from 2010} & 132 & 222\\ \\ 
  \shortstack{Average ESS of the REs,\\boundaries from 2010} & 125 & 245\\
      \bottomrule 
    \end{tabular}
\end{table}


\section{Discussion and Future work}\label{sec:discussion}
In this work we have demonstrated the applicability of aggregated GP priors to represent spatial random effect instead of traditional adjacency-based models, and presented a scalable solution to the change-of-support problem by jointly encoding GP aggregates using the PriorVAE technique. Modelling on fine resolution scales is attractive since this approach allows us to capture continuity, but it is computationally cumbersome. By introducing the aggVAE prior, we alleviate the computational difficulties. Our results showed that inference using aggVAE priors is orders of magnitude faster and more efficient than inference performed using the GP priors; effective sample size per second is thousands times higher when using aggVAE prior than combining the original GP priors and the aggregation step to obtain aggGP. Our work lays foundation for future extensions allowing to capture heterogeneity of continuous covariates $X$, such as environmental factors, at a fine spatial scale, by including them into the linear predictor of the model as the fixed effect term: $b_0 + X\beta + f$. We used the RBF kernel to model GP on the fine spatial scale. This kernel defines smooth and stationary GP draws. However, the presented methodology is kernel-agnostic, and any other kernel can be used instead, including non-stationary kernels.%
 One drawback of the PriorVAE method is pertinent in the current work as well: aggVAE is not explicitly encoding hyperparameters of the GP, such as lengthscale, and, hence, is not able to infer them. Future extensions should focus on closing this gap, e.g. conditional variational autoencoders can be used instead to overcome this issue \citep{semenova2023priorcvae}.
Since aggVAE provides a prior that does not have a closed form solution but is rather obtained in an empirical way by training a neural work, theoretical properties of such priors and their influence on downstream inference should be studied in more detail.%
While modelling prevalence, we have taken the number of positive and negative tests at their face values. Sensitivity of the test, however, may play a role. We also treated survey locations as noise-free, while due to privacy they have 10 km precision. Both facts should be taken into account while performing modelling for real-life applications and constitute future work.

\paragraph*{Data and Code Availability}
Data containing administrative boundaries of Kenya are publicly available: current boundaries\footnote{\url{https://data.humdata.org/dataset/2c0b7571-4bef-4347-9b81-b2174c13f9ef/resource/03df9cbb-0b4f-4f22-9eb7-3cbd0157fd3d/download/ken\_adm\_iebc\_20191031\_shp.zip}} and old boundaries\footnote{\url{https://www.wri.org/data/kenya-gis-data}} can be freely downloaded.
Malaria prevalence data was obtained from DHS 2015 survey and contains information on locations of clusters and test positivity to calculate district-specific prevalence; it can be requested from the DHS programme\footnote{\url{https://dhsprogram.com/}}. Code to reproduce the results is available at \url{https://github.com/MLGlobalHealth/aggVAE}.

\clearpage
\bibliography{uai2023-template}
\end{document}